\documentclass[runningheads]{llncs}
\usepackage{graphicx}
\usepackage{tikz}
\usepackage{comment}
\usepackage{amsmath,amssymb} 

\usepackage{color}
\usepackage{algorithm}
\usepackage{algorithmicx}
\usepackage{algpseudocode}

\usepackage{tikz}
\usepackage{comment}
\usepackage{amsmath,amssymb} 
\usepackage{graphicx}
\usepackage{color}
\usepackage{wrapfig}
\usepackage{booktabs}
\usepackage{cite}
\usepackage{multirow}
\usepackage[misc,geometry]{ifsym}
\usepackage[width=122mm,left=12mm,paperwidth=146mm,height=193mm,top=12mm,paperheight=217mm]{geometry}
\usepackage[accsupp]{axessibility}  

\usepackage[pagebackref,breaklinks,colorlinks]{hyperref}
\usepackage{blindtext}

\def\bez{B\'ezier\ }

\def\prf{\noindent{\bf Proof: }}
\def\eop{\rule{0.1in}{0.1in}\par\bigskip}

\begin{document}

\pagestyle{headings}
\mainmatter
\def\ECCVSubNumber{194}  

\title{ExtrudeNet: Unsupervised Inverse Sketch-and-Extrude for Shape Parsing } 
\titlerunning{ExtrudeNet}

\author{Daxuan Ren\inst{1,2} \and
Jianmin Zheng\textsuperscript \Letter, \inst{1} \and 
Jianfei Cai\inst{1,3} \and \\
Jiatong Li\inst{1,2} \and
Junzhe Zhang\inst{1,2}}
\authorrunning{D. Ren et al.}

\institute{Nanyang Technological University, Singapore \and
Sensetime Research \and
Monash University\\
\email{\{daxuan001, asjmzheng, E180176, junzhe001\}@ntu.edu.sg, jianfei.cai@monash.edu}}

\maketitle

\begin{abstract}
Sketch-and-extrude is a common and intuitive modeling process in computer aided design. This paper studies the problem of learning the shape given in the form of point clouds by ``inverse'' sketch-and-extrude. We present {\em ExtrudeNet}, an unsupervised end-to-end network for discovering sketch and extrude from point clouds. Behind ExtrudeNet are two new technical components: \textbf{1)} an effective representation for sketch and extrude, which can model extrusion with freeform sketches and conventional cylinder and box primitives as well; and \textbf{2)} a numerical method for computing the signed distance field which is used in the network learning.   
This is the first attempt that uses machine learning to reverse engineer the sketch-and-extrude modeling process of a shape in an unsupervised fashion. ExtrudeNet not only outputs a compact, editable and interpretable representation of the shape that can be seamlessly integrated into modern CAD software, but also aligns with the standard CAD modeling process facilitating various editing applications, which distinguishes our work from existing shape parsing research. Code is released at \textcolor{magenta}{\url{https://github.com/kimren227/ExtrudeNet}}.

\end{abstract}

\section{Introduction}

Pen draws a line, paint roller sweeps a surface, and pasta maker extrudes Fusilli from a stencil. From a point to a line, then to a surface and to a solid shape, the process of using lower dimensional shapes to construct a higher dimensional object seems to be a human instinct. In this paper, we explore the inverse of this process by training a neural network to infer 2D drawings of a point cloud and then extrude them into 3D to reconstruct the shape.
 
With recent development in 3D reconstruction technologies and cheaper sensors, point clouds can be easily obtained and become a widely adopted 3D data representation~\cite{qi2017pointnet, qi2017pointnet++, wang2019dynamic}. However, the unordered and unstructured nature of point clouds makes it difficult to perform high level manipulation and easy editing of their underlying geometries. Thus in recent years, the research of extracting shape features and generating high level shape representation from point clouds is very active, especially in computer vision and graphics.

\setlength{\columnsep}{10pt}
\setlength{\intextsep}{5pt}
\begin{wrapfigure}{r}{0.5\textwidth}
  \centering
  \includegraphics[width=.5\textwidth]{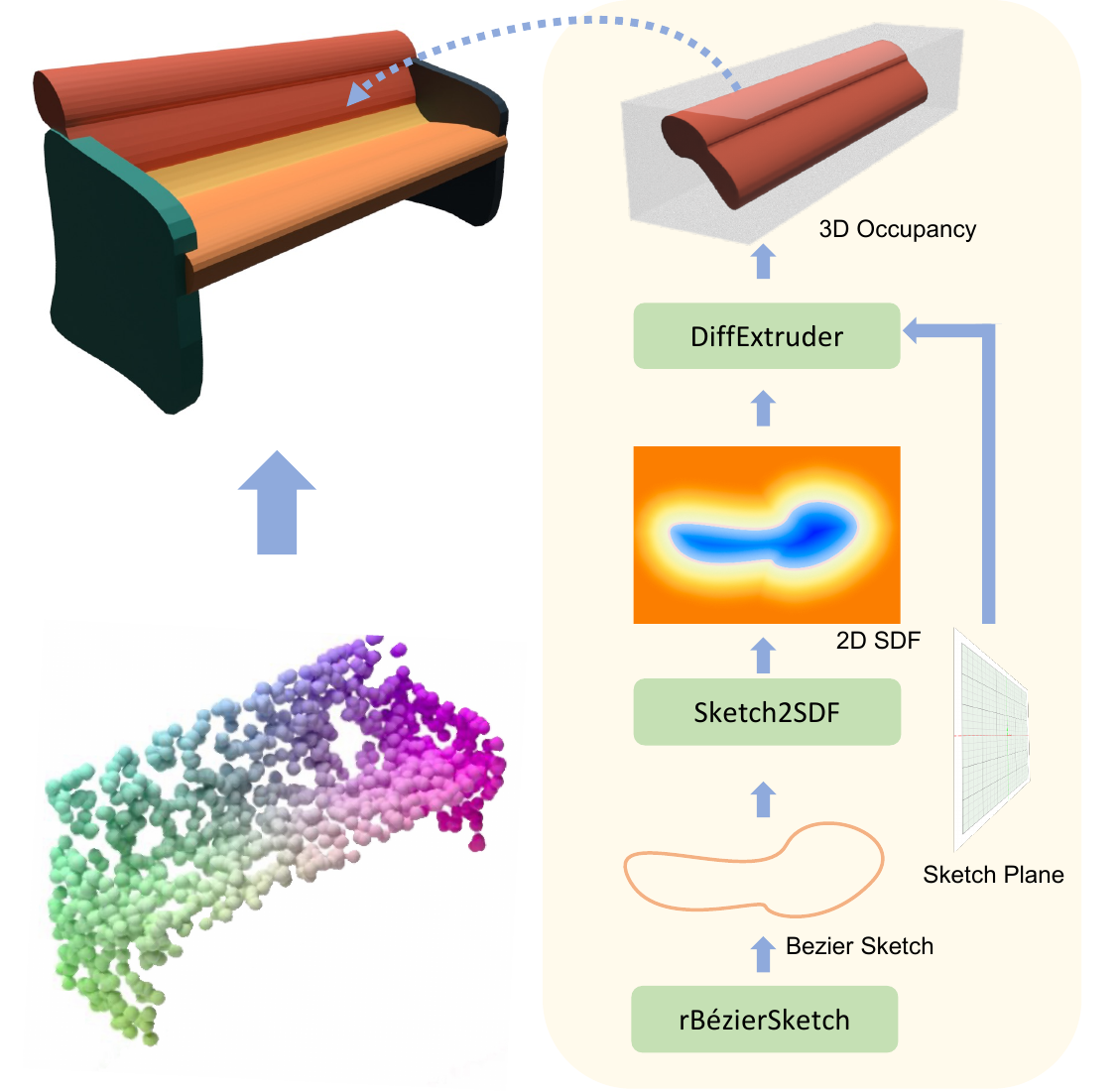}
  \caption{\small ExtrudeNet studies the problem of learning the shape, given in the form of point clouds, by ``inverse'' sketch-and-extrude.
  }
  \label{fig:teaser}
\end{wrapfigure}

We take inspiration from the process of sketch and extrude, a popular and intuitive approach widely used in the field of computer aided design (CAD) where engineers usually model shapes by first ``sketching'' a closed free form sketch (profile) in a 2D sketch plane and then ``extruding'' the sketch into 3D. 
We propose {\bf \em ExtrudeNet}, the first of its kind end-to-end unsupervised network for learning high level (editable and interpretable) shape representation through inverse sketch and extrude process from point clouds. 

To realize ExtrudeNet, as shown in Fig.~\ref{fig:teaser}, we create three modular components: {\em 1) rB\'ezierSketch}, which generates a simple closed curve (i.e.\ no self-intersection); {\em 2) Sketch2SDF}, a versatile numerical method for computing Signed-Distance-Field from parametric curves; and {\em 3) DiffExtruder}, a differentiable method for extruding 2D Signed-Distance-Field (SDF) into a 3D solid shape. Built upon these components, ExtrudeNet takes a point cloud as input and outputs sketch and extrude parameters which form a compact, interpretable and editable shape representation. ExtrudeNet's outputs are highly compatible with modern CAD software~\cite{fusion360}, allowing control points based editing, which is much easier compared to directly editing triangle, polygonal meshes or even primitive based constructive solid geometry (CSG) models~\cite{requicha1977constructive}.

There are prior works on converting point clouds to high level shape representations. These representatives are discovering CSG in either a supervised or unsupervised manner. 
Supervised approaches~\cite{sharma2018csgnet,wu2021deepcad,ganin2021computer,para2021sketchgen} suffer issues such as invalid syntax, infeasible models, and  requiring large amount of expert annotated data. Unsupervised methods~\cite{paschalidou2019superquadrics, tulsiani2017learning, chen2020bsp,deng2020cvxnet, ren2021csg, yu2021capri} find the Boolean combinations of pre-defined geometric primitives such as  box and cylinder. Our ExtrudeNet goes beyond these works. First, 2D sketch can be complex freeform curves, which allows us to model much more complex shapes using a single extrusion. Second, ``Sketch-and-Extrude'' is more user-friendly when it comes to editing and secondary-development, as editing a 2D sketch is more intuitive than editing 3D parameters. This Sketch-and-Extrude process happens to be a widely adopted method in CAD software for modeling 3D shapes ~\cite{fusion360,solidworks,kintel2014openscad,shapr3d}, making our method highly compatible with industry standards.  Moreover, extensive experiments show that our ExtrudeNet can reconstruct highly interpretable and editable representations from point clouds. We also show through qualitative visualizations and quantitative evaluation metrics that ExtrudeNet outputs better overall results.  The main contributions of the paper are:
\begin{itemize}
    \item We present an end-to-end network, ExtrudeNet, for unsupervised inverse sketch and extrude for shape parsing. To the best of our knowledge, ExtrudeNet is the first unsupervised learning method for discovering sketch and extrude from point clouds.
    \item We design a special rational cubic \bez curve based representation for sketch and extrusion learning, which can model freeform extrusion shapes, and the common cylinder and box primitives as well.
    \item We present a simple and general numerical method for computing the signed distance field of 2D parametric curves and their 3D extrusions which is proven to be suitable for gradient-based learning.
\end{itemize}

\section{Related Work}

\noindent \textbf{Shape Representation.}
There have been different representations for 3D shapes. Recently implicit representation ~\cite{mescheder2019occupancy, park2019deepsdf, hao2020dualsdf}, usually in the form of Occupancy or Signed Distance Field, has drawn a lot of attention. It frees from intricate data representation and can be used directly via neural networks. To extract the underlying geometry, however, further processing is required ~\cite{lorensen1987marching},  which is computationally intensive. Parametric representation  describes shapes by parametric equations and is widely used in industry for modeling shapes thanks to its strong edibility and infinite resolution. However,  generating parametric shapes from raw data like point clouds is a non-trivial task.

\noindent \textbf{High Level Shape Learning.}
High level shape representations are often required, which benefit various practical applications. With the advance in machine learning, learning high level shape representations from raw data structure gains popularity.
There have been many works for reconstructing CAD and especially CSG from point clouds. CSG is a tree-like structure representing shapes by starting from primitive objects and iteratively combining geometric shapes through Boolean operations~\cite{laidlaw1986constructive}.
It is well adapted in professional CAD software.  

CSGNet~\cite{sharma2018csgnet} pioneers supervised CSG learning by modeling CSG as a sequence of tokens. It processes the sequence into a valid CSG-Tree using NLP techniques. With recent NLP technologies, DeepCAD~\cite{wu2021deepcad} and CAD-As-Language~\cite{ganin2021computer} employ more powerful language models, e.g. Transformer, and add additional constraints to better predict CAD models. However, modeling CAD as language gives rise to addition problems (such as producing grammatically correct but invalid representations), which are not easy to solve. 

VP~\cite{tulsiani2017learning} pioneers the unsupervised approaches by using the union of a set of boxes to approximate shapes. SQ~\cite{paschalidou2019superquadrics} takes a step further by using super quadrics instead of boxes to better approximate complex shapes. BSPNet~\cite{chen2020bsp} and CVXNet~\cite{deng2020cvxnet} propose to use the union of a set of convexes to represent a complex shape, where the convexes are constructed by intersecting half-spaces. These methods extend the modeling capability of using standard primitives, but abundant planes are required to approximate freeform surfaces, which also limits their edibility.
UCSGNet~\cite{kania2020ucsg} proposes CSG-Layers that iteratively selects primitives and Boolean operations for reconstruction.
CSG-StumpNet~\cite{ren2021csg} reformulates CSG-Tree with arbitrary depth into a fixed three layer structure similar to Disjunctive Normal Form, and uses a simple network to generate binary matrices for select fundamental products to union into the final reconstruction. However, these methods focuses on CSG operations and use only basic primitives (boxes, spheres, etc.), which is not efficient to approximate complicated shapes. CAPRI-Net~\cite{yu2021capri} uses quadric implicit shapes to construct two intermediate shapes using an approach similar to BSP-Net. The two intermediate shapes are then subtracted to form the final shape. The construction process is interpretable, but the use of quadric implicit shapes reduces its edibility.

\begin{figure*}[htp]
  \centering
  \includegraphics[width=1\textwidth]{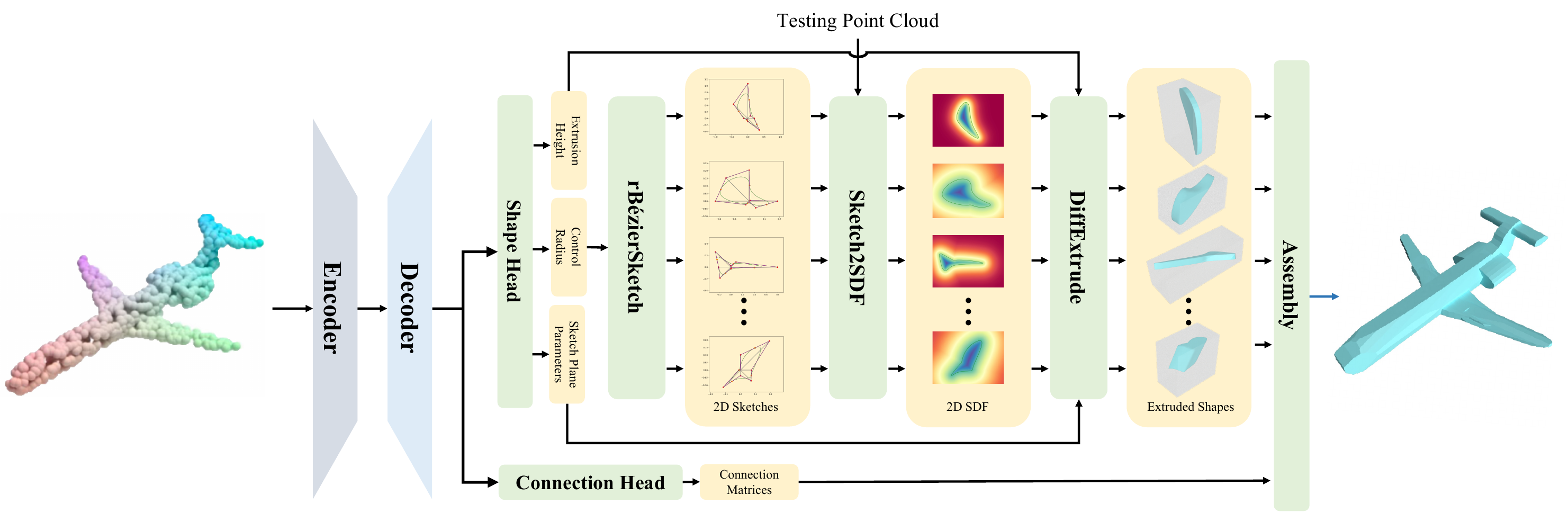}
  \caption{\textbf{Framework overview.} The input point cloud first goes through the encoder-decoder phase to predict shape parameters and connection matrices via shape head and connection head. rB\'ezierSketch is then used to generate the profile curve from the decoded sketch parameters. Sketch2SDF computes the SDF of the sketches. Sketch SDF, sketch plane parameters, and extrusion height are passed into DiffExtruder to generate the occupancy of the extrusion shapes, which together with the connection matrices are passed into CSG-Stump to generate the final shape occupancy.}
  \label{pipeline}
\end{figure*}
\section{ExtrudeNet}
This section presents ExtrudeNet, an end-to-end network for unsupervised inverse sketch and extrude for compact and editable shape parsing. The input to the network is a point cloud representing a shape to be learned. ExtrudeNet outputs a set of 3D extrusions as the building blocks and their Boolean operations, which together create the shape. Each of the 3D extrusions is defined by a 2D sketch profile curve and a 3D extrusion process.

The pipeline of the entire network is illustrated in Fig.~\ref{pipeline}. The main components are briefly described below. Note that different encoder and assembly methods can be used to adapt for different use cases.
\begin{enumerate}
    \item[(1)] {\bf Encoder-Decoder:} ExtrudeNet first encodes the input point cloud into a latent feature using the off-the-shelf DGCNN as a backbone encoder~\cite{wang2019dynamic}.  The latent code is then enhanced by three fully connected layers with size 512, 1024, and 2048. After that, the latent feature is passed to the shape head and the connection head to decode into extruded shape parameters and connection matrices. Extruded shape parameters consist of 2D sketch parameters, sketch plane parameters, and extrusion height. Connection matrices represent Boolean operations among the extruded shapes. Since binary value is not differentiable, we use the Sigmoid function to predict a soft connection weight in $[0, 1]$ for each matrix.
    \item[(2)] {\bf rB\'ezierSketch:} rB\'ezierSketch is used to convert sketch parameters into a closed profile curve defined by a set of rational cubic B\'{e}zier curves for extrusion. The generated 2D sketch curve is guaranteed to be free of self-intersection and can represent free-form curves, circular arc and even polygon in a single formulation.
    \item[(3)] {\bf SDF-Generation:}  This consists of Sketch2SDF and DiffExtruder. Sketch2SDF is first used to compute the Signed-Distance-Field (SDF) of the generated 2D sketch on a plane. The computed 2D sketch SDF, the sketch plane parameters, and extrusion height are then passed into DiffExtruder to compute the extruded shapes' occupancy field in 3D space. 
    \item[(4)] {\bf Assembly:} 
    Given the predicted extrusion shapes' occupancy and connection matrices, we are in a position to assemble the extrusion shapes to complete the final reconstruction. For this purpose, we choose to directly use CSG-Stump from ~\cite{ren2021csg} for its simplicity and learning friendly nature. CSG-Stump reformulates CSG-Trees into a three layer structure similar to Disjunctive Normal Form and use three fixed size binary matrices to generate and select fundamental products. The first layer of CSG-Stump is a complement layer indicating whether the complement of the input occupancy should be used for the down steam operations. The second layer is an intersection layer which selects and intersects complement layer's outputs into intersected shapes. The last layer selects intersected shapes to union into the final shape.
\end{enumerate}
Below we describe rB\'ezierSketch and SDF-Generation in more detail.

\subsection{rB\'ezierSketch and Extrusion}
To create an extrusion shape, a profile curve should be sketched and then extruded in 3D space.
We use a network to predict the sketch parameters that define the profile curve in the $XY-$plane (serving as its local coordinate system) and then extrude it along the $Z-$direction to create the extrusion shape. The shape is then transformed to the required location and orientation predicted by the network which mimics the sketch plane transform in CAD software.
\subsubsection{rB\'ezierSketch.}
Considering the common modeling practice and shapes in CAD applications, there are a few assumptions for the profile curve: (i) it is closed and has no self-intersection, which has advantages in defining a valid solid shape; (ii) it is piecewise smooth for creating quality shapes; and (iii) it can model freeform curves, and circles or polygons as well. Meanwhile, we also have to balance the capability and complexity of the representation such that it can be easily deployed into a learning pipeline. Based on these considerations, we propose the following model for our profile curve.

The basic mathematical model is a closed curve formed by $N$ curve segments defined by special rational cubic B\'{e}zier curves $C_k(t), t\in [0,1], k=0,1,\cdots,N-1$, which may explain the name {\em rB\'ezierSketch}. 
The equation of $C_k(t)$ is:
\begin{equation}
    C_k(t) = \frac{P^k_0B_0^3(t)+ w^k_1 P^k_1B_1^3(t)+ w^k_2 P^k_2B_2^3(t)+ P^k_3B_3^3(t)}{B_0^3(t)+w^k_1 B_1^3(t)+ w^k_2B_2^3(t)+B_3^3(t)}
\label{eq:rBezier}
\end{equation}
where  $P^k_i = (x_i^k,y_i^k)$ are the control points on the $XY-$plane, weights $w_1^k\ge 0, w_2^k\ge 0$ are for the two inner control points, and $B_i^3(t) = {3\choose i}(1-t)^{3-i}t^i$ are Bernstein polynomials. To make sure that consecutive segments are connected to form a closed curve, the constraints $P^k_3 = P^{(k+1)\text{mod} N}_0$ are added.

The closed curve is defined around the origin. Thus it is convenient to express each control points $P^k_i = (x^k_i, y^k_i) = (\rho_i^k\cos(\alpha_i^k), \rho^k_i\sin(\alpha^k_i))$ by
the radial coordinate $\rho^k_i$  and  the polar angle $\alpha_i^k$.  For simplicity, we further distribute the central angles of the segments evenly. That is, each B\'{e}zier curve has the central angle $\frac{2\pi}{N}$. Within each segment $C_k(t)$, the polar angles of control points are chosen to be:
\begin{equation}
    \alpha^k_1= \alpha^k_0+\theta,\;\; \alpha^k_3 = \alpha^k_0+\frac{2\pi}{N},\;\; \alpha^k_2 = \alpha^k_3-\theta
\label{eq:polarAng}
\end{equation}
where $\displaystyle \theta =\frac{2\pi}{4N}+\tan^{-1}\left(\frac{1}{3}\tan\left(\frac{2\pi}{4N}\right)\right)$. It is worth pointing out that these polar angles are specially designed to achieve the capability of circle recovery (see Proposition~2 below). In this way, to specify the B\'{e}zier control points, we just need to provide the radial coordinates. Connecting all the control points in order forms a polygon that is homeomorphic to the origin-centered unit circle, which assures good behavior of the generated profile curve. In summary, the network only has to estimate the radial coordinates $\rho^k_i$ and the weights $w^k_1, w^k_2$ in order to sketch the profile curve.

\noindent
{\bf Remark 1.} The proposed curve model is specially designed to deliver a few nice properties, which are outlined in the following paragraphs.

Rational cubic B\'{e}zier representation in (\ref{eq:rBezier}) is proposed because it is a simple form of NURBS that is the industry standard in CAD and meanwhile sufficient to model freeform curves~\cite{farin01}. Besides freeform smooth shapes, this representation is able to represent a straight line or a polygon, for example, as long as we let all the control points $P^k_i$ lie on a line. The reason that the rational form is chosen is that it includes polynomial curves as a special case and has the capability of exactly representing a circle. These properties enable the extrusion shapes to include box and cylinder primitive shapes as special cases.

Due to the special angular set-up of rB\'ezierSketch, the generated profile curve is a simple closed curve, i.e., it does not self-intersect. In fact, we have an even stronger result.
\begin{proposition}
The area bounded by the curve generated by rB\'ezierSketch is a star-shaped set.
\label{proposition:star-shaped-set}
\end{proposition}

\begin{wrapfigure}{r}{0.5\textwidth}
  \centering
  \includegraphics[width=.5\textwidth]{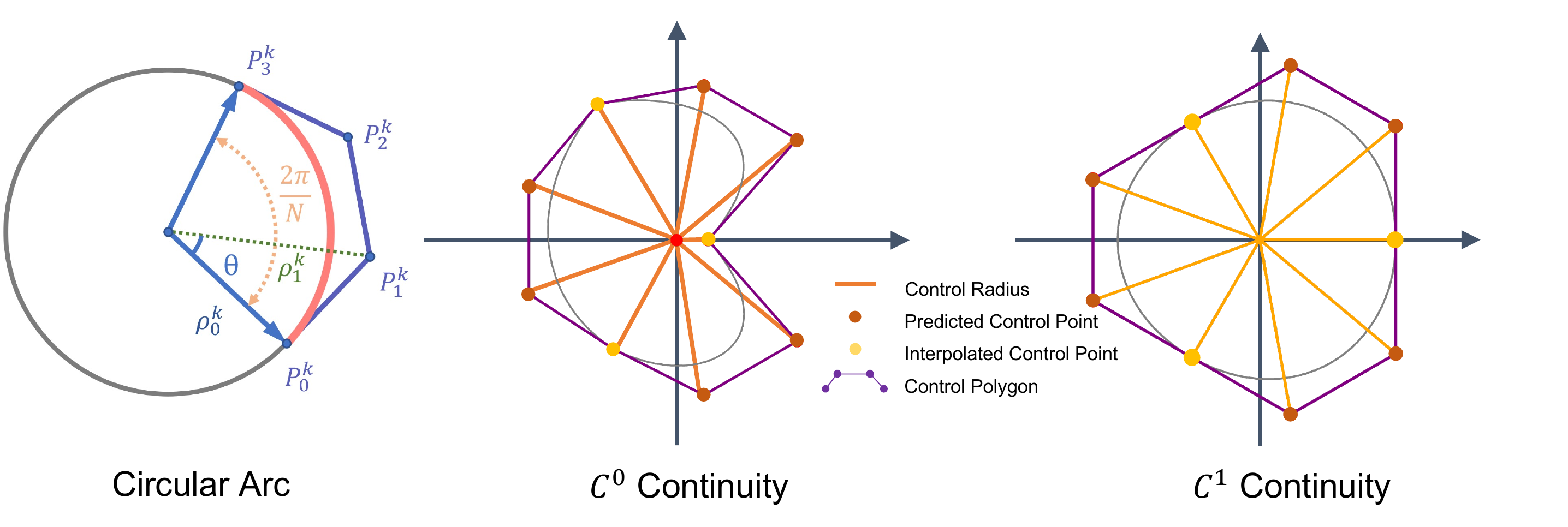}
  \caption{\small Left: a circular arc. Middle: a $C^0$ profile curve. Right: a $C^1$ profile curve.
  }
  \label{fig:circular_arc_c0_c1}
\end{wrapfigure}

Moreover, the special choice of angle $\theta$ in (\ref{eq:polarAng}) makes it possible to exactly represent a circular arc using Eq.~\ref{eq:rBezier}. 
\begin{proposition} Let the polar angles be given by Eq.~\ref{eq:polarAng}. If $\rho_0^k = \rho_3^k$, $\rho_1^k = \rho_2^k = \frac{\rho^k_0}{\cos(\theta)}$, and $w^k_1 = w^k_2 = \frac{1}{3}\left(1+2 \cos\left(\frac{\pi}{N}\right)\right)$,
then the rational \bez curve of (\ref{eq:rBezier}) defines a circular arc, as shown in Fig.~\ref{fig:circular_arc_c0_c1} (left).
\end{proposition}

The proposed curve model assures $C^0$ continuity among the segments since the \bez control polygons are connected in a loop by enforcing each curve's last control point to be the same as the next curve's first control point (see Fig.~\ref{fig:circular_arc_c0_c1} (middle)). Thus totally $3N$ radial coordinates and $2N$ weights need to be predicted. In case the predicted control points and weights happen to satisfy certain condition given by Proposition~3, $C^1$ continuity can be achieved.
\begin{proposition}
If the control points and weights satisfy 
\begin{equation}
    P_0^{k+1} = P_3^k = \frac{\rho_2^{k}P^{k+1}_1+\rho^{k+1}_1P^k_2}{\rho_2^{k}+\rho_1^{k+1}}, \; \frac{w^k_2}{w^{k+1}_1} = \frac{\rho^{k+1}_1}{\rho^k_2},
    \label{eq:c1}
\end{equation}
curve segments $C_{k+1}(t)$ and $C_k(t)$ meet at $P_0^{k+1}$ with $C^1$ continuity.
\end{proposition}

\noindent The proof of above propositions is provided in the supplementary material.

\setlength{\columnsep}{10pt}
\setlength{\intextsep}{5pt}

If for some CAD applications we already have prior knowledge that the models should be at least $C^1$ continuous, then we can enforce $C^1$ continuity by letting $P^k_0$ and $w_2^k$ be computed from Eq.~\ref{eq:c1}. In this case we just predict $2N$ control points $P^k_1, P^k_2$ and $N$ weights $w_2^k$, which have fewer variables (see Fig.~\ref{fig:circular_arc_c0_c1} (right)). 

\noindent
{\bf Remark 2.} In CAD there have been some works to define single-valued curves in the polar coordinate system. For example, Sanchez-Reyes proposed a subset of rational \bez curves that can be used to define single-valued curves~\cite{sanchez1990single} and later extended them to splines~\cite{sanchez1992single}. It should be pointed out that our proposed curves are different from those proposed by Sanchez-Reyes. Particularly, for a Sanchez-Reyes's curve, the control points are on radial directions regularly spaced by a constant angle and each weight must equal the inverse of the radial coordinate of the corresponding control point, which leaves very few degrees of freedom (DoF) for shape modeling. Our curves have more DoFs.
\subsubsection{Extrusion.}
Once the profile curve $C(t) = (x(t), y(t))$ on the $XY-$plane is obtained, the extrusion shape is generated by directly extruding the curve along the $Z-$direction. The extrusion shape is bounded by the top and bottom planes and a side surface. The side surface has the parametric equations
$(x,y,z) = ((x(t),y(t),s)$ where $s\in [0,h]$ is the second parameter and $h$ is the extrusion height estimated by the network.

\begin{wrapfigure}{R}{0.5\textwidth}
  \centering
  \includegraphics[width=.5\textwidth]{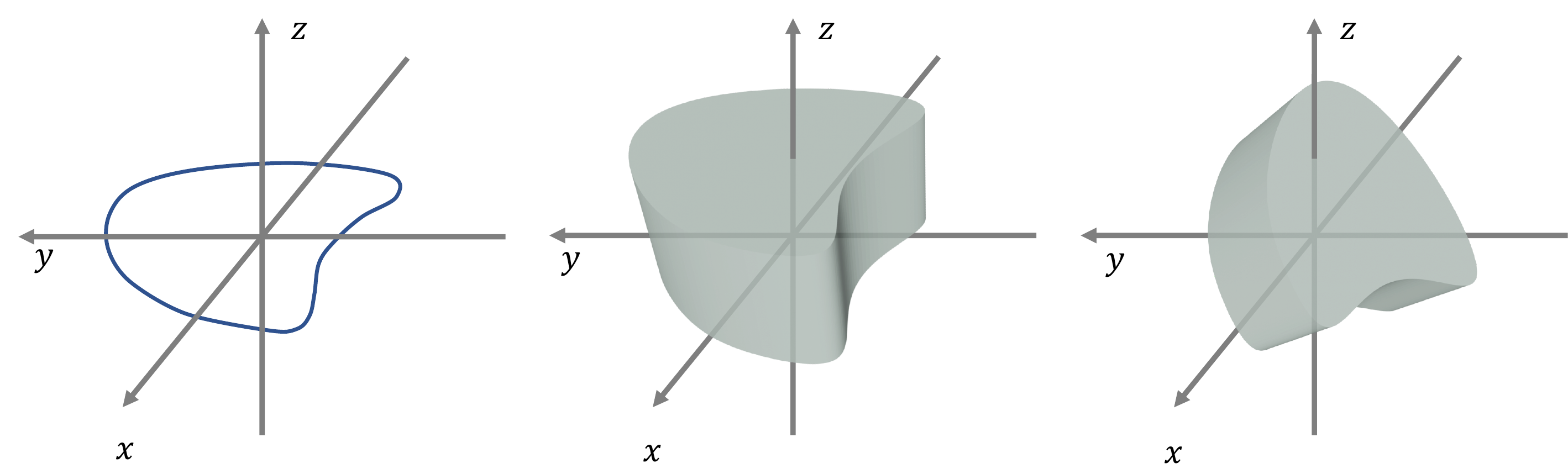}
  \caption{\small Left: 2D sketch in the $XY-$plane; Middle: direct extrusion; Right: the target extrusion shape.}
  \label{fig:extrusion}
\end{wrapfigure}

The network also estimates a quaternion that defines a rotation matrix $R$ and a translation vector ${\bf t} = (t_x, t_y, t_z)$ that defines a translation matrix $T$. The matrix $R$ makes the $XY-$plane be in the orientation of the target sketch plane and the matrix $T$ moves the origin to the target sketch center on the sketch plane. Thus applying matrices $R$ and $T$ to the upright extrusion shape gives the target extrusion shape in 3D space, as shown in Fig.~\ref{fig:extrusion}.   
\setlength{\columnsep}{10pt}
\setlength{\intextsep}{5pt}

\subsection{SDF-Generation}
\subsubsection{Sketch2SDF.}
\label{Sketch2SDF} 
Note that the generated sketches are in parametric forms. Unlike explicit or implicit functions, computing a signed-distance-field of parametric curves is not trivial. There were a few implicitization algorithms for converting a parametric curve into an implicit representation~\cite{sederberg1984implicit,ZhengS01}. However, they require exact arithmetic computation. Moreover, as observed in \cite{SederbergZKD99}, the existence of singularity in parametric representation often makes the resulting implicit expression useless.  

\setlength{\columnsep}{10pt}
\setlength{\intextsep}{5pt}
\begin{wrapfigure}{r}{0.5\textwidth}
  \centering
  \includegraphics[width=.5\textwidth]{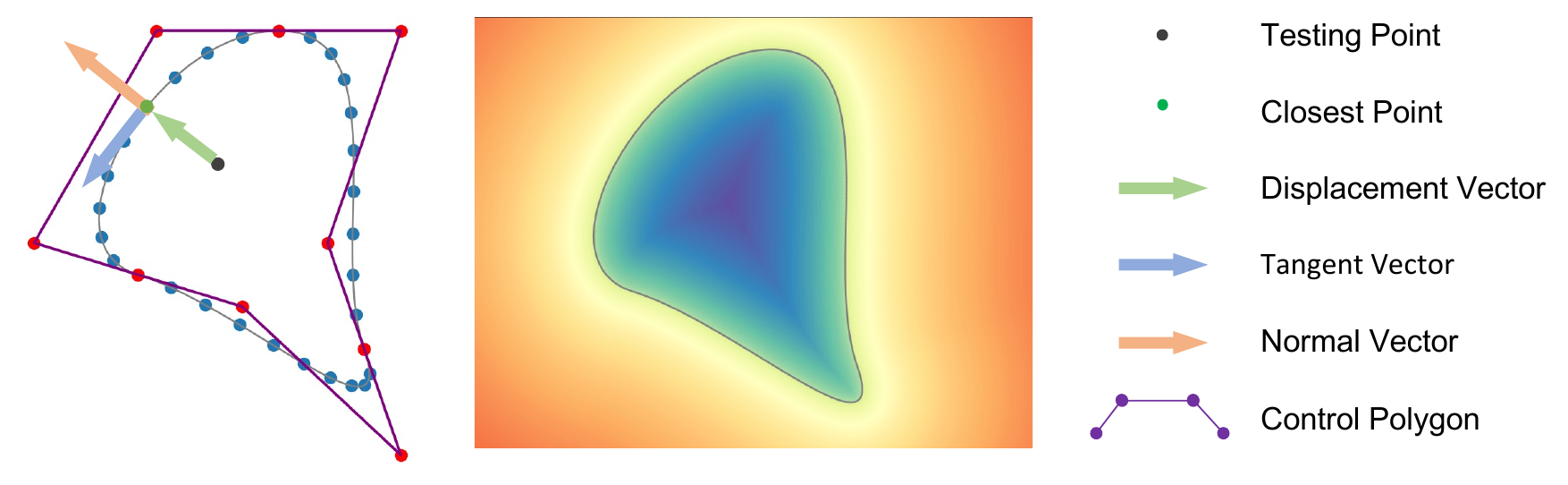}
  \caption{\small The SDF of a curve is computed by a numerical method.}
  \label{fg:SDF2D_sketchs}
\end{wrapfigure}

In this section, we present a numerical method for computing the SDF of parametric sketches. The method is general.
While it applies to the rational cubic \bez curves here, it also works for other parametric curves. This method also yields good gradients which is friendly for deep learning applications (see Sec.~\ref{2DExperiment}). The SDF of a sketch is defined by the Distance-Field $DF(p)$, the smallest distance from a given testing point $p$ to the curve, multiplied by a sign $SIGN(p)$ which indicates whether the testing point $p$ is inside or outside of the sketch. 

\noindent\textbf{Distance-Field.}
To compute the Distance-Field of a sketch, we first sample a set of points $S$ from the curve and take the smallest distance between the test point $p$ and sample points $S$ as the value of $DF(p)$. Specifically, given a parametric curve $C(t)= (x(t), y(t)), t\in[0,1]$, the sampling points are obtained by evenly sampling the parameter values in the parameter domain:
\begin{equation}
\label{sampling}
S =\left\{\left(x(\frac{i}{n}),y(\frac{i}{n})\right)\mid i=0,1,\cdots,n\right\}
\end{equation}
Then the distance between the testing point $p$ and the set $S$ is found by
\begin{equation}
\label{distancefield}
DF(p)= \underset{s\in S}{\min} \|s-p\|_2
\end{equation}
and its corresponding closest point's parameter value is denoted by $CT(p)$.

\noindent\textbf{Signed Distance-Field.}
To test if a point $p$ is inside or outside of a sketch, we check whether if the displacement vector from the testing point towards the closest point in $S$ has the same direction as the normal vector of the curve at the closest point.
We assume that the curves are parameterized such that they are traced in the counter-clockwise direction. Otherwise, a simple reparameterization can correct it. Then
the normal vector of the curve $C(t)$ can be computed by rotating its tangent vector by 90 degrees clockwise:
\begin{equation}
\label{normal}
N(t) =  \left(\frac{d y(t)}{dt}, -\frac{d x(t)}{dt}\right)
\end{equation}
If two consecutive curve segments of the sketch meet with $C^0$ continuity, the normal vectors of the two segments at the joint point are different. To ensure a consistent normal at the junction of two curves, the average of the two is used. 
In this way, the sign function for a testing point $p$ can be computed:
\begin{equation}
\label{sign}
SIGN(p)= \frac{N(CT(p)))\cdot (C(CT(p))-p)}{\|N(CT(p)))\cdot (C(CT(p))-p)\|_2 + \epsilon}
\end{equation}
where a small positive number $\epsilon$ is added to prevent from zero division.
Finally, the Signed-Distance of $p$ is:
\begin{equation}
\label{SDFofSketch}
SDF_s(p) = SIGN(p) \times DF(p). 
\end{equation}
Fig.~\ref{fg:SDF2D_sketchs} gives an example of a computed SDF.

\subsubsection{DiffExtruder.}
\label{DiffExtrude}
Now we show how to compute the SDF of the extrusion shape. We first transform the testing point $p$ back to a point $p'$ by reversing the transformations that transform the $XY-$plane to the target sketch plane: $p' = R^{-1}(T^{-1}(p))$. Then we compute the signed distance of $p'$ with respect to the upright extrusion shape whose base lies on the $XY-$plane. For this purpose, we project point $p'$ onto the $XY-$plane by setting its z-value to 0 and denote the footprint by $p^{\prime\prime}$, and further find its nearest point $(c_x, c_y, 0)$ on the profile curve. Then the projected testing point $p^{\prime\prime}$, the nearest point $(c_x,c_y,0)$ and its corresponding point $(c_x,c_y,h)$ on the end plane of the extrusion define a vertical plane. Now, the problem of finding the SDF of the extrusion shape is reduced into computing the 2D SDF on the vertical plane. 

If a point is inside of the extrusion shape, its signed distance to the shape simply becomes the minimum of the 2D sketch SDF and the distances from the point to the two planes that bound the extrusion.
\begin{equation}
\label{sdf_inside}
    SDF_i(p') = \max(\min(SDF_s((p'_x, p'_y)), h-p'_z, p'_z), 0)
\end{equation}

If the point lies outside the extruded primitive, its signed distance becomes:

\begin{equation}
    \begin{array}{lcl}
     SDF_o(p') &=& -[
     (\min(h-p'_z,0))^2+(\min(p'_z,0))^2 \\
     \mbox{} & & +(\min(SDF_s((p'_x, p'_y)),0))^2 ]^{\frac{1}{2}}.
    \end{array}
\label{sdf_outside}
\end{equation}
It can be seen that
for a testing point outside the extruded shape, $SDF_i$ becomes zero, and for a testing point inside the extrusion shape, $SDF_o$ becomes zero. Therefore we can simply get the overall SDF by adding the two terms:
\begin{equation}
\label{sdf3d}
    SDF(p') = SDF_i(p') + SDF_o(p').
\end{equation}

\noindent\textbf{Occupancy of Extrusion Shape.}
After getting the SDF of an extrusion shape, the occupancy can be computed by a sigmoid function $\Phi$ where $\eta$ indicates how sharp the conversion is taken place:
\begin{equation}
\label{sdf2occ}
    O(p') = \Phi(-\eta\times SDF(p')).
\end{equation}    

\subsection{Training and Inference}
We train ExtrudeNet in an end-to-end and unsupervised fashion as no ground truth sketch and extrusion parameters are present. The supervision signals are mainly generated by the reconstruction loss $L_{re}$ that computes the discrepancy between the reconstructed shape's occupancy $\hat{O}$ and the ground truth shape's occupancy $O^*$ using a set of testing points $P\subset {R^3}$ sampled within the shape's bounding box:
\begin{equation}
L_{re} = \mathbb{E}_{p\sim P}||\hat{O}_i-O^*_i||_2^2. 
\end{equation}
As suggested in ~\cite{ren2021csg}, when a shape is too far from any testing points, the gradient becomes extremely small due to the use of a sigmoid function. Thus a primitive loss is introduced to ``poll'' the primitive shapes towards testing points: 
\begin{equation}
L_{prim} = \frac{1}{K}\sum_k^K \min_{n}SDF_{k}^2(p_n),
\end{equation}

We enforce the weights to be close to one, which encourage the rational \bez curves to be close to \bez curves.

Thus, the overall objective is then defined: 
\begin{equation}
L_{total} = L_{re} + \lambda_{p} \cdot L_{prim}  + \lambda_{w} \cdot \sum\limits_{k=0}^{N-1}\sum\limits_{i=1}^2(w^k_i-1)^2 
\end{equation}
where $\lambda_{p}$ and $\lambda_{w}$ are the trade-off factors.

During inference, we discretize the connection matrices into Boolean values for interpretable construction. In addition, we implement a CAD converter that directly takes binary connection matrices and the network generated parameters as input and converts the reconstructed shape into CAD compatible format. We use OpenSCAD ~\cite{kintel2014openscad} to render the final shape into STL~\cite{roscoe1988stereolithography} format.

\section{Experiments}

In this section, we first evaluate our complete pipeline ExtrudeNet, particularly its ability to extract editable and interpretable shape representations with ablation on different settings. Then, we evaluate the key component of rB\'ezierSketch and Sketch2SDF on 2D toy examples.

\begin{figure*}[t]
  \centering
  \includegraphics[width=1\textwidth]{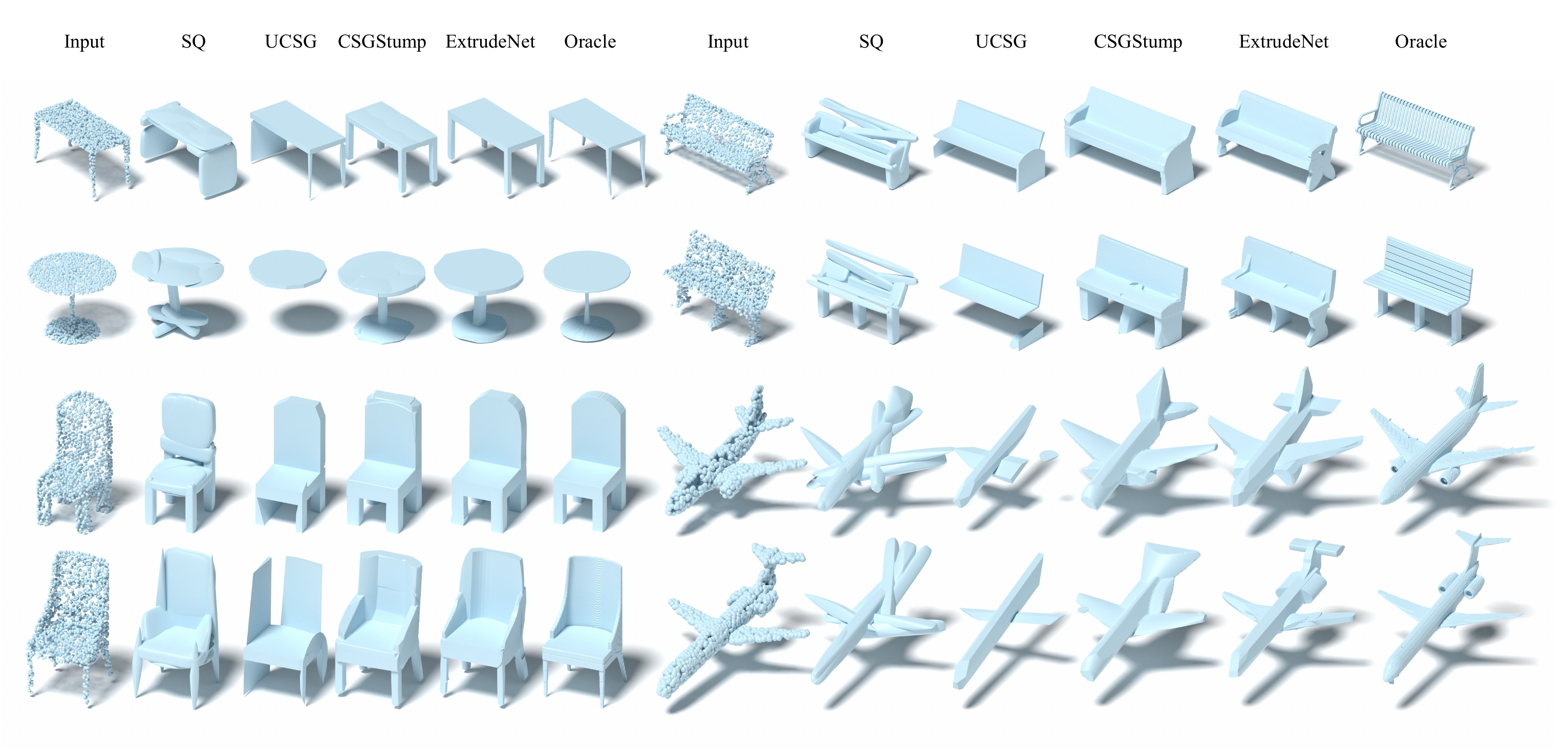}
  \caption{Qualitative comparison between ExtrudeNet and other baselines. ExtrudeNet generates visually more pleasing outputs with curved surfaces compared to all the baselines.}
  \label{fg:baseline_compares} 
\end{figure*}

\subsection{Evaluations of ExtrudeNet}
\noindent
\textbf{Dataset.}
We evaluate the ExtrudeNet on ShapeNet Dataset ~\cite{chang2015shapenet} with train, test, and val split aligned with prior methods. The input point clouds are sampled from the original mesh surface via Poisson Disk Sampling\cite{bridson2007fast}. The testing points are sampled using~\cite{battysdfgen} in a grid from the mesh bounding box with 15\% of padding on each side. This padding is important to remove unwanted artifacts, see Supplementary Material.

\noindent
\textbf{Implementation Details.}
We implement ExtrudeNet using PyTorch~\cite{NEURIPS2019_9015}, and train the network with Adam Optimizer~\cite{kingma2014adam} with a learning rate of 1e-4. We train each class on a single NVIDIA V100 GPU with a batch size of 16. It took about 5 days to converge.

In our experiments, each sketch is constructed with 4 B\'{e}zier curves with a sample rate of 100. We estimate 64 extruded shapes in total, and use 64 as the number of intersection nodes in Assembly (see Fig.~2) adopted from CSG-Stump~\cite{ren2021csg}. 

\noindent \textbf{Main Results - Comparisons with Baselines.}
We compare our method with related approaches focusing on editable shape abstraction, namely, VP ~\cite{tulsiani2017learning}, SQ~\cite{paschalidou2019superquadrics}, UCSG-Net~\cite{kania2020ucsg}, and CSG-Stump Net~\cite{ren2021csg}. Note that BSPNet and CVXNet are excluded as they mainly focus on the reconstruction of thousands of planes that contain too many primitives and thus lack editability. For fair comparison, we align the number of available primitives to 64 for VP, SQ and UCSG. We show qualitative comparisons with the baseline methods in Fig.~\ref{fg:baseline_compares}. It can be seen that our ExtrudeNet models curved surfaces more effectively compared to all the baseline methods, thus resulting in a more detailed reconstruction. 

Moreover, we also conduct quantitative evaluations using symmetric $L_2$ Chamfer Distance (CD), volumetric IoU and surface F1 scores. The comparisons of the results are given in Table~\ref{tab:compare_baseline}. We can see that ExtrudeNet achieves the best overall reconstruction quality.

\begin{table}[ht]
\caption{Different Metric computed between 3D reconstruction results and the ground truth shapes. ExtrudeNet outputs better overall results}
\begin{center}

\resizebox{\textwidth}{!}{
\begin{tabular}{c |ccc |ccc |ccc |ccc |ccc}

\multirow{2}{*}{}
&\multicolumn{3}{c|}{\textbf{VP~\cite{tulsiani2017learning}}}
&\multicolumn{3}{c|}{\textbf{SQ~\cite{paschalidou2019superquadrics}}}
&\multicolumn{3}{c|}{\textbf{UCSG ~\cite{kania2020ucsg}}}
&\multicolumn{3}{c|}{\textbf{Stump~\cite{ren2021csg}}}
&\multicolumn{3}{c}{\textbf{ExtrudeNet}}\\  
 & CD  &  V-IoU  & F1  &  CD  &  V-IoU  & F1  & CD  &  V-IoU  & F1  & CD  & V-IoU  & F1  & CD  &  V-IoU  & F1  \\
\hline
Chair       &\textbf{1.10}	&0.28	&\textbf{67.07}	&2.85  	&0.21  	&41.49 	&3.54	&0.25	&61.01	&1.34	&\textbf{0.41}	& 63.33	&1.47	&\textbf{0.41}	&60.17 \\
Car         &1.02	&\textbf{0.67}	&62.71	&1.25	&0.17	&78.58	&\textbf{0.64}	&0.11	&\textbf{82.54}	&0.76	&0.32	&76.45	&0.67	&0.41	&82.20 \\
Sofa        &2.18	&0.29	&48.75	&1.27	&0.37	&69.14	&1.30	&0.29	&79.59	&0.85	&0.61	&\textbf{83.88}	&\textbf{0.79}	&\textbf{0.62}	&83.42 \\
Plane       &5.11	&0.29	&37.55	&\textbf{0.58}	&0.23	&82.94	&0.71	&0.10	&\textbf{87.66}	&0.70	&\textbf{0.36}	&74.57	&0.66	&0.33	&77.95 \\
Lamp        &8.41	&\textbf{0.24}	&35.57	&\textbf{1.79}	&0.16	&\textbf{63.00}	&7.02	&0.22	&55.66	&3.48	&\textbf{0.24}	&57.93	&3.77	&0.20	&42.86 \\
Telephone   &2.75	&0.55	&48.22	&0.46	&0.38	&88.29	&\textbf{0.34}	&\textbf{0.69}	&\textbf{93.32}	&1.60	&0.58	&91.24	&0.56	&0.62	&89.00 \\
Vessel      &2.84	&0.37	&54.21	&\textbf{0.76}	&0.27	&\textbf{78.52}	&4.54	&0.09	&61.65	&1.15	&0.44	&72.30	&1.63	&\textbf{0.48}	&61.06 \\
Loudspeaker &1.67	&0.45	&50.06	&2.07	&0.34	&65.00	&1.81	&0.19	&56.35	&1.70	&0.52	&63.32	&\textbf{1.21}	&\textbf{0.60}	&\textbf{75.04} \\
Cabinate    &3.16	&0.48	&42.90	&1.97 	& 0.31 	&40.52	&1.09	&0.38	&73.56	&0.77	&0.56	&79.11	&\textbf{0.73}	&\textbf{0.68}	&\textbf{84.94} \\
Table       &1.62	&0.26	&60.05	&2.89	&0.17	&60.01	&3.64	&0.30	&65.09	&1.35	&0.35	&\textbf{74.48}	&\textbf{1.06}	&\textbf{0.45}	&73.49 \\
Display     &1.25	&0.36	&60.52	&\textbf{0.72}	&0.33	&80.09	&1.13	&0.54	&78.18	&1.64	&0.50	&75.75	&0.96	&\textbf{0.55}	&\textbf{83.66} \\
Bench       &1.57	&0.26	&63.31	&1.09	&0.17	&73.46	&1.73	&0.21	&74.12	&1.04	&0.29	&73.40	&\textbf{0.78}	&\textbf{0.39}	&\textbf{81.17} \\
Rifle       &1.35 	&0.35	&66.06 	&\textbf{0.38}	&0.26	&\textbf{90.55}	&1.05	&0.29	&84.38	&0.78	&0.37	&85.61	&0.76	&\textbf{0.44}	&84.18 \\

\hline  
\textbf{Mean} & 2.61  &0.37   &53.61	&1.39 	&0.25  	&70.12  &2.19  	&0.28  	&73.31  & 1.32  &0.42  	& 74.64 & \textbf{1.15} 	& \textbf{0.47} & \textbf{75.32} 
\end{tabular}
}

\end{center}
\label{tab:compare_baseline}
\end{table}

\setlength{\columnsep}{10pt}
\setlength{\intextsep}{4pt}
\begin{wraptable}{r}{0.52\textwidth}
\centering
\caption{CD results ($\times10^{-3}$) with shapes extruded from different numbers of sketches.
}

\begin{tabular}{c c c c c }
\hline
\#Primitives & 8 & 16 &32 & 64 \\
 \hline
Plane & 0.89 &0.96 &0.81 &\textbf{0.66} \\
Chair& 2.01 &1.81  &1.47 &\textbf{1.47}  \\
Bench& 1.41 &1.18  &1.30 &\textbf{0.78} \\
\hline
\end{tabular}
\label{tab:num-primitive}

\end{wraptable}

\noindent \textbf{Effect of Different Numbers of Primitives.}
To show that our proposed extrude shapes with freeform profile curves are highly adaptable, we train ExtrudeNet with limited numbers of extruded shapes. Table~\ref{tab:num-primitive} reports the results on the three classes.
It can be seen that even with only 8 available extruded shapes, our method can still achieve reasonable reconstruction results.

\setlength{\columnsep}{10pt}
\setlength{\intextsep}{5pt}
\begin{wraptable}{r}{0.52\textwidth}
\setlength{\tabcolsep}{4pt}
\centering
\caption{CD results ($\times10^{-3}$) with shapes extruded from different sketches on the Airplane class.
}

\begin{tabular}{c c c c c}
\hline
Sketch Type & Freeform & Circle & Poly \\
\hline 
CD & \textbf{0.66} & 0.877 & 0.838 \\
\hline
\end{tabular}
\label{tab:diff-curves}

\end{wraptable}

\noindent \textbf{Effect of Different Sketches.} 
Our method can also use other profile curves such as polygonal and circular. 
Table~\ref{tab:diff-curves} gives a comparison on the CD results under different types of profile curves. We can see that extruded shapes with the proposed freeform profile curves outperform the results using the other two common profile curves by a significant margin, which indicates the strong representability of the extruded shapes with freeform profile curves.

\begin{figure}[h]
  \centering
  \includegraphics[width=0.9\textwidth]{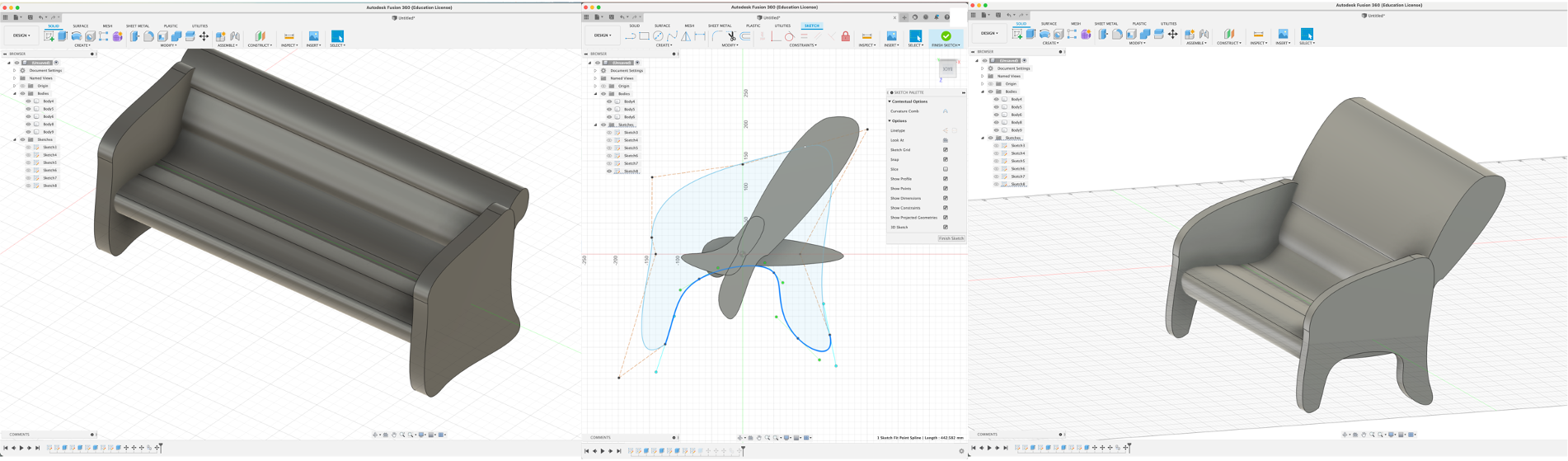}
  \caption{As ExtrudeNet's output is directly compatible with Industrial CAD software, we can directly import the shape into Fusion360 and edit using its GUI interface. We show that a bench can be edited into an armchair by simply edit the 2D cross-section sketch by dragging the control points and adjust the extrusion height. \textbf{Left:} ExtrudeNet's output. \textbf{Middle:} Edit sketch using Fusion 360. \textbf{Right:} Result after editing.}\label{fig:edit}

\end{figure}

\noindent \textbf{Edibility of Sketch and Extrude.}
After getting the results from ExtrudeNet, a designer can do a secondary development with ease. Using the bench example in Fig.~\ref{fig:edit}, we show that by editing the Bezier Control Points and reducing the extrusion height, we can easily generate an armchair from the reconstruction result.

\noindent
More ablations such as sampling rate, number of curves, sketch plane placement etc., can be found in the supplementary material.
 
\subsection{Evaluation of rB\'ezierSketch}\label{2DExperiment}
\begin{figure}[h]
  \centering
  \includegraphics[width=\textwidth]{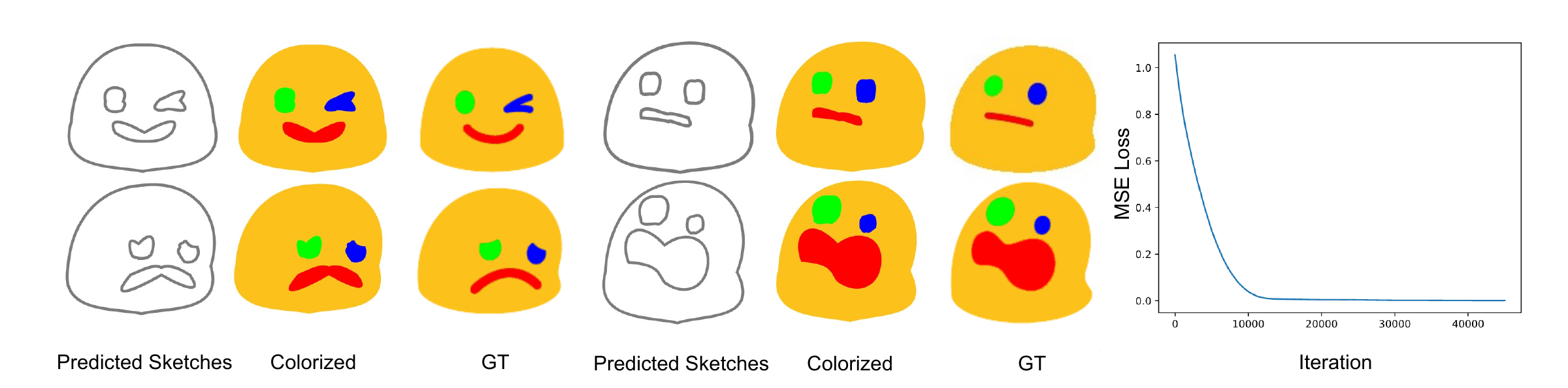}
  \caption{Left: We show that using rB\'ezierSketch alone we can reconstruct the emojis with good qualities. Right: Our rB\'ezierSketch and Sketch2SDF provide low variance gradients which yields a smooth convergence. }\label{fig:2d_exp}
\end{figure}

To demonstrate the approximation ability and the ease of learning of the proposed rB\'ezierSketch, we conduct a fitting experiment that directly optimize for the radial coordinate that best reconstructs a given raster emoji image from~\cite{notoEmoji}. As the emojis contains different color blocks, which is not suitable to be represented using occupancy or SDFs, we choose to approximate the individual boundaries instead of solid colors blocks. Given a raster emoji image, we first compute its edges using Canny edge detection,  
and then we compute the Distance Field for the edge image. We fit the sketch by minimizing the MSE loss between the predicted and ground truth distance fields using standard gradient decent. As shown in Fig.~\ref{fig:2d_exp}, the emoji can be reconstructed by rB\'ezierSketch with good quality. We also show the training plot, where a smooth decrease in the loss term indicating that rB\'ezierSketch and Sketch2Field yields good gradients and it is friendly to used in a learning setup.

\subsection{Evaluation of Sketch2SDF}
Here we demonstrate the versatility and accuracy of Sketch2SDF on computing parametric sketches' SDF by giving concrete examples. 

\begin{figure}[ht] 
	\begin{minipage}{0.4\textwidth}
	
	\includegraphics[width=\linewidth]{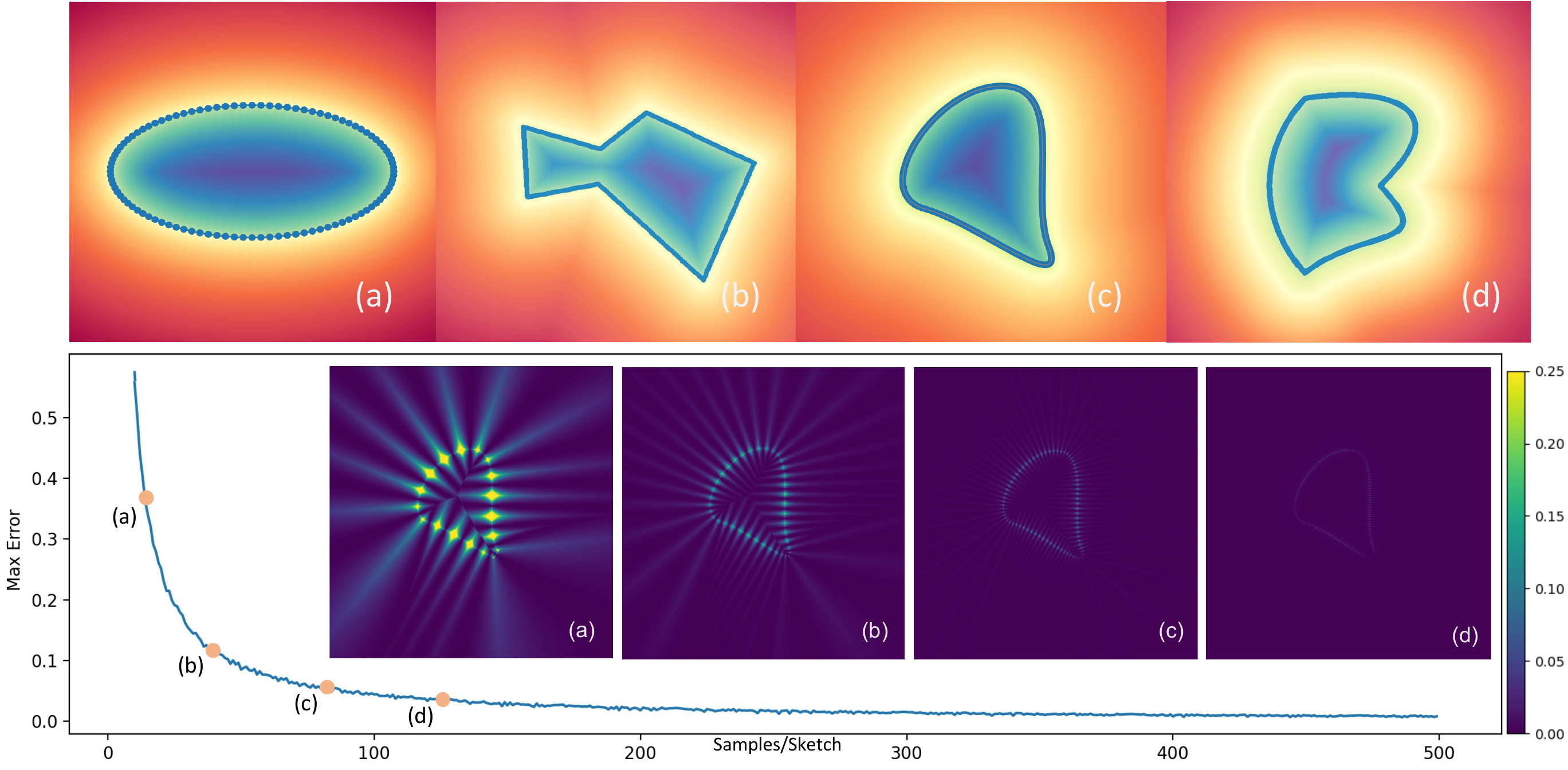}
	\end{minipage}
	\begin{minipage}{0.59\textwidth}
		\caption{Top: Sketch2SDF can be easily adapted to different parametric curve types.
		Bottom: Visualization of the approximation errors by comparing numerical and analytical SDFs of Bezier sketch. (a)$\sim$(d) indicate the errors at sampling rate of 20, 40, 80 and 120. We can see that error is barely noticeable with a relatively small sampling rate.}
    \label{fg:sdf_error}
	\end{minipage}
\end{figure}

\noindent \textbf{Versatility.}
To show the versatility, we implemented four different kinds of parametric curves, namely polygon, ellipse, Cubic Bezier Sketchs with $C^1$ and $C^0$ continuity, see Fig.~\ref{fg:sdf_error}(Top). Note that our method is not limited to these curves, and can be easily adapted to new parametric curve types. 
During the implementation of these curves, only the curve sampling and normal computation need to be adapted accordingly while the rest of the code remains untouched. This indicates that our pipeline is highly reusable and can support new parametric curves with minimum effort.

\noindent \textbf{Approximation Accuracy.}
As Sketch2SDF is a numerical method, approximation is introduced, where the sample rate is the most influential factor. Thus, we study the degree of approximation under different sampling rates by comparing our numerical results against the analytical results using B\'{e}zier Sketches.
In Fig.\ref{fg:sdf_error}, we plot the maximum error of all testing points versus sample rate and visualize the error heat map. 
We can see that too few sample points will dramatically decrease the accuracy, and higher sampling rates yield more accurate approximation, which is however at the cost of higher computation and memory usage. Empirically, we find that around 400 sample points per curve is a reasonable trade-off.
\section{Conclusion}
In this paper, we have presented ExtrudeNet, which is an effective framework for
unsupervised inverse sketch-and-extrude for shape parsing from point clouds. It not only output a compact, editable and interpretable shape representation but also a meaningful sketch-and-extrude process, which can benefit various applications. As demonstrated by extensive experiments, ExtudeNet can express complex shapes with high compactness and interpretability while achieving state-of-the-art reconstruction results both visually and quantitatively. 

\noindent \textbf{Discussion.}
Currently, ExtrudeNet requires a relatively long training time as sketches are more freeformed and have higher degrees of freedom. We also notice that some artifacts have been reconstructed due to high degrees of freedom. The current extrusion is a simple extrusion along the direction orthogonal to the sketch plane. Generalizing the extrusion to include more advanced processes such as sweeping and lofting warrants further investigation, which would have more impacts in the industry.

\noindent \textbf{Acknowledgements}
The work is partially supported by a joint WASP/NTU project (04INS000440C130), Monash FIT Startup Grant, and SenseTime Gift Fund.

\clearpage

\bibliographystyle{splncs04}
\bibliography{egbib}
\clearpage
\begin{center}
  {\Large \bf {ExtrudeNet: Unsupervised Inverse Sketch and Extrude for Shape Parsing} \\
-- Supplementary Materials --
}
 \end{center}
\def\thesection{\Alph{section}}
\def\thefigure{\Alph{figure}}
\def\thetable{\Alph{table}}
\section{Propositions and Proofs}
\begin{proposition}
The area bounded by the curve generated by rB\'ezierSketch is a star-shaped set.
\end{proposition}
\prf
Consider an arbitrary ray cast from the origin. Since the polygon generated by connecting all the control points in order is homeomorphic to the origin-centered unit circle, the ray will intersect the polygon at least once. On the other hand, based on the construction of the control points of the rational \bez curve, when we travel along the polygon in the counter-clock direction, the polar angle is monotonically increasing, which means that the ray will not intersect the polygon more than once. Therefore the ray intersects the \bez control polygon only once, so does it intersect the profile curve, which is confirmed by the continuity of the curve and the variation diminishing property of rational \bez curves. The variation diminishing property says that for an arbitrary line, the number of intersections with the curve will not be greater than the number of intersections with the control polygon~\cite{farin01}. Hence any point on the profile curve can be visible to the origin. \eop 

\begin{proposition} Let the polar angles be given by Eq.2. If $\rho_0^k = \rho_3^k$, $\rho_1^k = \rho_2^k = \frac{\rho^k_0}{\cos(\theta)}$, and $w^k_1 = w^k_2 = \frac{1}{3}\left(1+2 \cos\left(\frac{\pi}{N}\right)\right)$,
then the rational \bez curve of (1) defines a circular arc, as shown in Fig.3 (left).
\end{proposition}
\prf
Consider triangle $\triangle OP_0^kP_1^k$ shown in Fig.3 (left). If $\rho_1^k  = \frac{\rho^k_0}{\cos(\theta)}$, $\angle OP_0^kP_1^k$ is a right angle. Then $\|P_0^kP_1^k\| = \rho^k_0 \tan(\theta)$. Since
$\displaystyle \theta =\frac{\pi}{2N}+\tan^{-1}\left(\frac{1}{3}\tan\left(\frac{\pi}{2N}\right)\right)$, we thus have
\[
\begin{array}{lcl}
\|P_0^kP_1^k\| &=&   \displaystyle
\rho_0^k \frac{\tan\left(\frac{\pi}{2N}\right)+\frac{1}{3}\tan\left(\frac{\pi}{2N}\right)}{1-\tan\left(\frac{\pi}{2N}\right)\cdot \frac{1}{3}\tan\left(\frac{\pi}{2N}\right)} \\
\mbox{} & = & 
\displaystyle
\rho_0^k \frac{\frac{4}{3} \frac{\sin\left(\frac{\pi}{2N}\right)}{\cos\left(\frac{\pi}{2N}\right)}}{1-\frac{1}{3}
(\sec^2\left(\frac{\pi}{2N}\right)-1)} 
\\
\mbox{} & = & \displaystyle 
\rho_0^k \frac{4\sin\left(\frac{\pi}{2N}\right) \cos\left(\frac{\pi}{2N}\right)}{4\cos^2\left(\frac{\pi}{2N}\right)-1} \\
\mbox{} & =& \displaystyle
\rho_0^k \frac{2\sin\left(\frac{\pi}{N}\right)}{1+2\cos\left(\frac{\pi}{N}\right)}.
\end{array}
\]
Similar analysis applies to the other end of the \bez curve. It can be easily verified that these equations together with the chosen weights in the proposition make the conditions given in \cite{WANG1991283} be satisfied, which ensure that the generated rational cubic \bez curve is an circular arc, and thus we arrive at the conclusion. \eop

\begin{proposition}
If the control points and weights satisfy 
\begin{equation}
    P_0^{k+1} = P_3^k = \frac{\rho_2^{k}P^{k+1}_1+\rho^{k+1}_1P^k_2}{\rho_2^{k}+\rho_1^{k+1}}, \; \frac{w^k_2}{w^{k+1}_1} = \frac{\rho^{k+1}_1}{\rho^k_2},
    \tag{S.1}
    \label{eq:c1}
\end{equation}
curve segments $C_{k+1}(t)$ and $C_k(t)$ meet at $P_0^{k+1}$ with $C^1$ continuity.
\end{proposition}
\prf
Consider curve segments $C_k(t)$ and $C_{k+1}(t)$. First, $P_0^{k+1} = P_3^k$ assures $C^0$ continuity of the two segments. Second, from Eq.S.1, we have
\begin{equation}
    \rho_2^k(P_1^{k+1}-P_0^{k+1}) = \rho_1^{k+1} (P^k_3 -P^k_2)
    \tag{S.2}
    \label{eq:bisector}
\end{equation}
or 
\begin{equation}
    w_1^{k+1}(P_1^{k+1}-P_0^{k+1}) = w_2^{k} (P^k_3 -P^k_2).
    \tag{S.3}
    \label{eq:c1condition}
\end{equation}

Eq.S.2 implied that $P_0^{k+1}$ or $P_3^k$ does lie on the bisector of angle $\angle P_2^kOP_1^{k+1}$. Eq.S.3 means $C^{\prime}_k(1) = C^{\prime}_{k+1}(0)$. This is because
\[
  C^{\prime}_{k} (1) = 3 \left((P_3^{k}-w^k_2P_2^{k})-(1-w^{k}_2)P^{k}_3\right) 
   = 3 w_2^{k}(P_3^{k}-P^{k}_2)
\]
and similarly,
\[
  C^{\prime}_{k+1}(0) 
  = 3 w_1^{k+1}(P_1^{k+1}-P^{k+1}_0).
\]
Therefore $C_{k+1}(t)$ and $C_k(t)$ meet at $P_0^{k+1}$ with $C^1$ continuity. \eop

\section{More Ablation Studies}
\paragraph{Ablation on Number of Sample Points Per Curve}

As discussed in Fig.10, sample rate plays an important role on the accuracy of the approximation. To further demonstrate that Sketch2SDF can work with a relative small sampling rate, we vary the sampling rate and train ExtrudeNet separately, as shown in the Tab.\ref{tab:num_sample_points}, even with a small sample rate (80 per sketch) ExtrudeNet still manages to generate comparable results.
\begin{table}[h]
  \centering
  \caption{Chamfer Distances of Airplane class under different sample rate. We can see that ExtrudeNet performs comparably even under small number of sample points.} 
  \label{tab:num_sample_points}
  \begin{tabular}{c c c c c}
  \hline
  Sample Rate & 400 & 320 & 160 & 80 \\
  \hline      
  CD & \textbf{0.664} & 0.702 & 0.726 & 0.725 \\
  \hline  
  \end{tabular}
\end{table}

\paragraph{Ablation on Number of Bezier Curve Per Sketch}
rB\'ezierSketch can be configured to generate sketches with different number of Bezier Curves. Highly flexible sketch can be achieved via using a large number of curves, however, this increase the computation cost and also gives network too much degree of freedom, making the network much harder to train.

\begin{table}[h]
  \centering
  \caption{Performance of ExtrudeNet under different numbers of Curves. We can see that too few \bez curves is not flexible enough while large number of curve makes network harder to train.}
  \label{tab:num_curves}
  \begin{tabular}{c c c c c}
  \hline
  \# Curves   & 2 & 4 & 6 & 8 \\
  \hline      
  CD & 0.754 & \textbf{0.664} & 0.722 & 1.09 \\
  \hline  
  \end{tabular}
\end{table}

\paragraph{Ablation on Sketch Plane Orientation}
Three-view-drawing is a popular drawing method that placing the sketch plane on 3D coordinate planes. In this ablation, we study the effect of the ExtrudeNet's performance under constraint sketch plane orientation, see Tab.\ref{tab:sketch-plane}. Interestingly, the results are consistent with how human tends to draw shapes (from top and front).

\begin{figure}[t]
  \centering
  \includegraphics[width=0.7\textwidth]{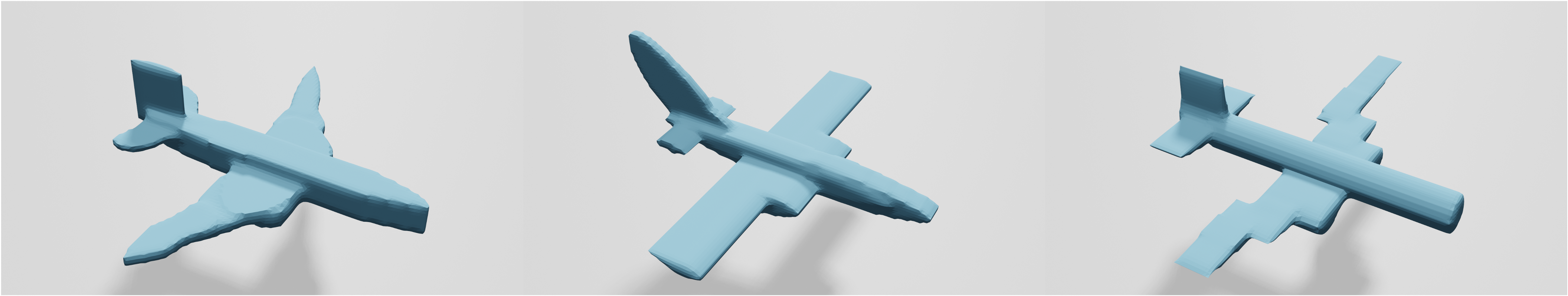}
  \caption{Reconstruction results by setting constraints on the extrusion plane orientation. \textbf{Left}: Top Extrude. \textbf{Middle}: Side Extrude. \textbf{Left}: Front Extrude.}
  \label{extrusion_plane}
\end{figure}

\begin{table}[h]
  \centering
  \caption{We train ExtrudeNet with constraint sketch plane orientation, we can see that freely placed sketch plane yields the best result.} 
  \label{tab:sketch-plane}
  \begin{tabular}{c c c c c}
  \hline
  Orientation   & side & top & front & free \\
  \hline      
  CD & 3.291 & 0.821 & 0.733 & \textbf{0.664} \\
  \hline  
  \end{tabular}
\end{table}

\setlength{\columnsep}{10pt}
\setlength{\intextsep}{5pt}
\begin{wrapfigure}{R}{0.4\textwidth}
  \centering
  \includegraphics[width=.4\textwidth]{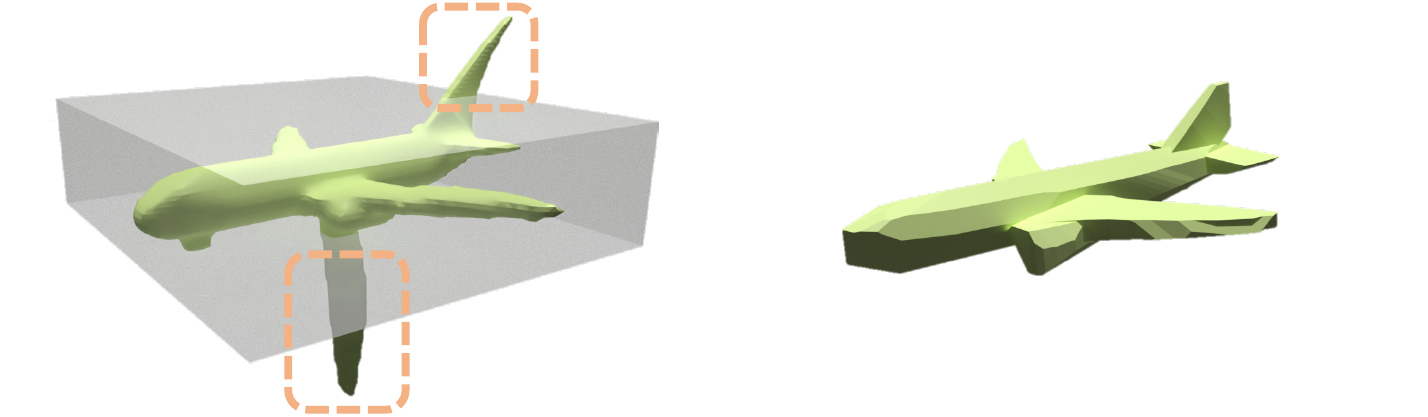}
  \caption{\small Left: Without padding the generated mesh contains artifacts that overshoot the gray bounding box. Right: Result with padding.
  }
  \label{fig:padding}
\end{wrapfigure}

\noindent \textbf{Effect of Padding.}
As \bez Sketches are very flexible and locally supported, only the control points that have testing points in their proximity are updated. Note that we add 15\% padding on each side of the mesh bounding box. Without padding,  artifacts that ``over-shoot" outside the bounding box can be generated (Fig.\ref{fig:padding}). Adding padding when sampling the testing points can solve this problem, as ``over-shot" shapes in the padded region will incur a higher loss which forces all the control points to be updated.

\section{More Visualizations}
\begin{figure}[h]
  \centering
  \includegraphics[width=1\textwidth]{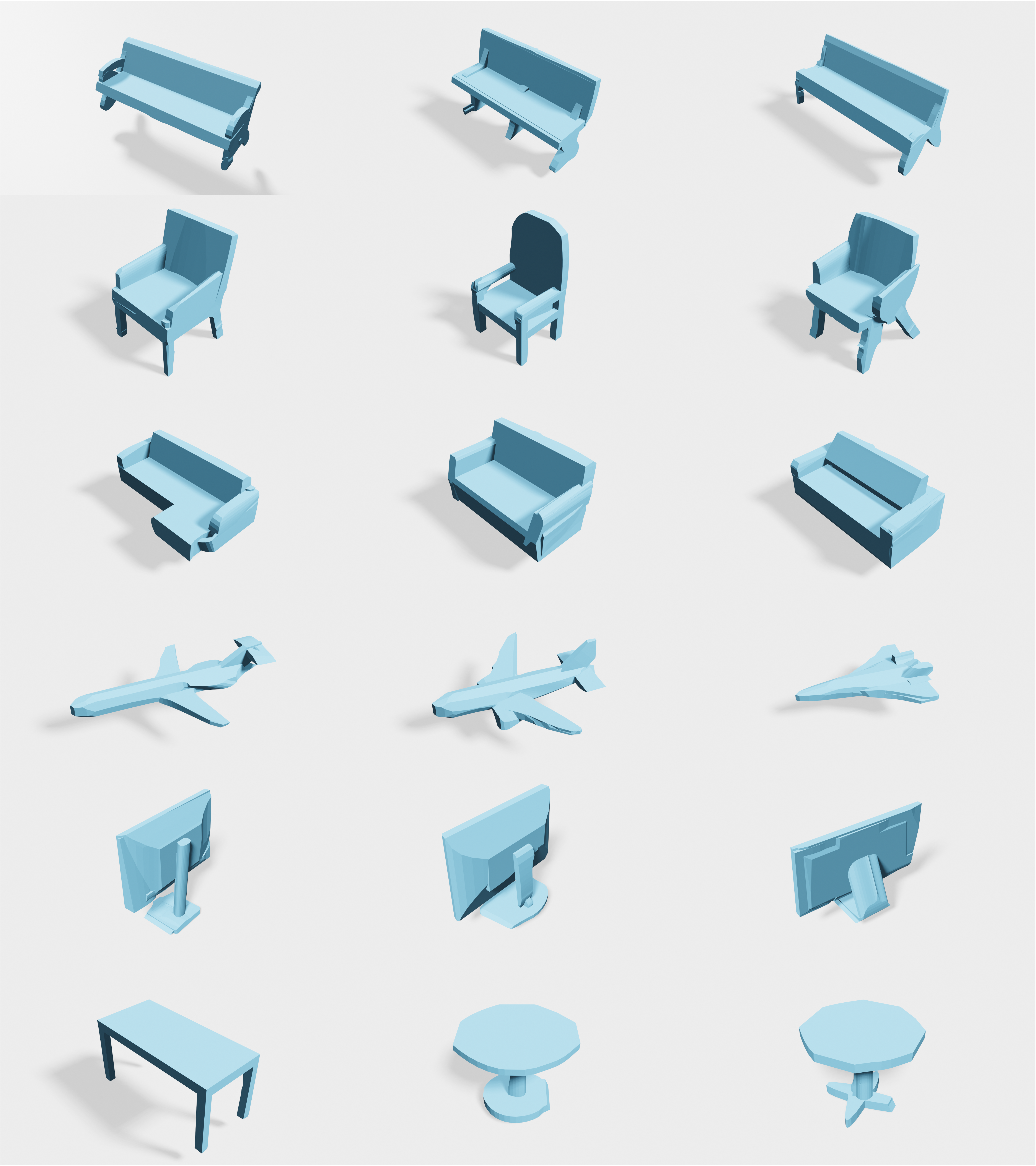}
  \caption{More visualization of ExtrudeNet results}
  \label{more_results}
  \vspace{-0.1in}
\end{figure}

We have created a video for better visualization, see attached. The video is composed of two sections. First section illustrates ExtrudeNet in detail with special attention on the "Sketch" and "Extrude" process. Note that, to better visualize individual extruded shapes, we using UNION as the assembly method, thus the final reconstruction results are worse than the complete ExtrudeNet. Second section of the video shows detailed reconstructed results. We have also attached more rendered results in Fig.\ref{more_results}. 

\end{document}